\documentclass[10pt,twocolumn,letterpaper]{article}

\usepackage[pagenumbers]{cvpr} 
\usepackage{amsmath}
\usepackage{mathpazo}
\usepackage{amssymb}
\usepackage{amsthm}
\usepackage{bm}
\usepackage{multirow}
\usepackage{subcaption,graphicx}
\usepackage{xspace}
\usepackage{mathtools}

\newcommand{\umin}{u_{\text{min}}}
\newcommand{\umax}{u_{\text{max}}}
\newcommand{\vmin}{v_{\text{min}}}
\newcommand{\vmax}{v_{\text{max}}}
\newcommand{\modelname}{NeuroNURBS\xspace}
\newcommand{\R}{\ensuremath{\mathbb{R}\xspace}}

\definecolor{cvprblue}{rgb}{0.21,0.49,0.74}
\usepackage[pagebackref,breaklinks,colorlinks,allcolors=cvprblue]{hyperref}

\title{NeuroNURBS: Learning Efficient Surface Representations for 3D Solids}

\author{Jiajie Fan\textsuperscript{1,2}\quad
Babak Gholami\textsuperscript{1} \quad
Thomas B\"ack\textsuperscript{2}\quad
Hao Wang\textsuperscript{2}\\
\textsuperscript{1} BMW Group, Bremer Str.~6, Munich, Germany\\
\textsuperscript{2} LIACS, Leiden University, Niels Bohrweg 1, Leiden, The Netherlands\\
\tt\small{\{jiajie.fan, babak.gholami\}@bmw.de} \quad
\tt\small{\{t.h.w.baeck,h.wang\}@liacs.leidenuniv.nl}
}

\begin{document}
\twocolumn[{
\maketitle
\vspace{-35pt}
\begin{center}
    \centering
    \captionsetup{type=figure}
    \includegraphics[width=0.95\linewidth, trim=0mm 15mm 0mm 0mm, clip]{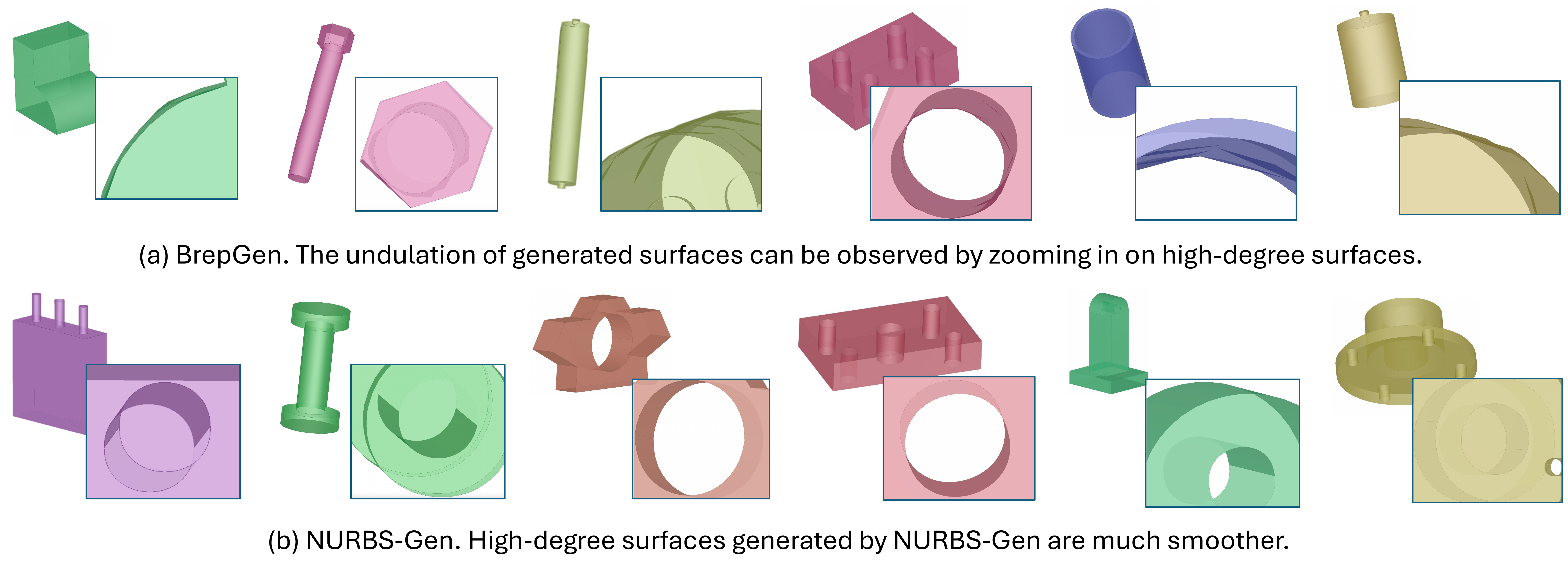}
    \captionof{figure}{Visual comparison of surfaces generated by BrepGen (a) and our method \emph{NURBS-Gen} (b). Our \emph{NURBS-Gen} can directly generate valid NURBS parametrization and hence ensure the smoothness and regularity of the surfaces, in contrast to the BrepGen surfaces, which appear undulating. 
    }
    \label{fig:Nurbs-Gen_qualitative_zoom}
\end{center}
}]
\begin{abstract}
    Boundary Representation (B-Rep) is the de facto representation of 3D solids in Computer-Aided Design (CAD). B-Rep solids are defined with a set of NURBS (Non-Uniform Rational B-Splines) surfaces forming a closed volume. To represent a surface, current works often employ the UV-grid approximation, i.e., sample points uniformly on the surface. However, the UV-grid method is not efficient in surface representation and sometimes lacks precision and regularity.  In this work, we propose NeuroNURBS, a representation learning method to directly encode the parameters of NURBS surfaces. Our evaluation in solid generation and segmentation tasks indicates that the NeuroNURBS performs comparably and, in some cases, superior to UV-grids, but with a significantly improved efficiency: for training the surface autoencoder, GPU consumption is reduced by 86.7\%; memory requirement drops by 79.9\% for storing 3D solids. Moreover, adapting BrepGen for solid generation with our NeuroNURBS improves the FID from 30.04 to 27.24, and resolves the undulating issue in generated surfaces.
\end{abstract}    
\section{Introduction}
\label{sec:intro}
\begin{figure*}[th]
    \centering
    \includegraphics[width=1\linewidth, trim=10mm 2mm 10mm 3mm, clip]{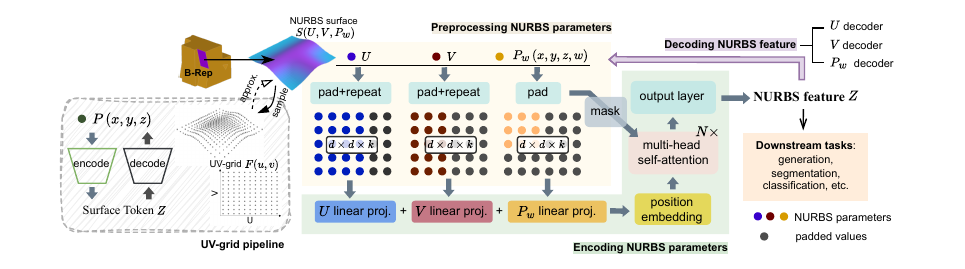}
    \caption{Diagram for \modelname. \textit{Right:} two parts of \modelname, preprocessing and autoencoding NURBS parameters. \textit{Left:} a simplified diagram for UV-grid, where the approximation from UV-grid back to NURBS surface is not deterministic.}
    
    \label{fig:NurbsEmb-Diagram}
\end{figure*}
Boundary Representation (B-Rep)~\cite{Weiler1986TopologicalSF,Lee2001PartialES} is commonly used to represent shapes and solids in computer-aided design (CAD) - widely applied in industrial design, simulation, and manufacturing. In a B-Rep, the solid boundaries are defined using a set of surfaces~\cite{Weiler1986TopologicalSF, nurbs_diff_2022}, which are, by default, parameterized by Non-Uniform Rational B-Splines (NURBS)~\cite{piegl_nurbs_1996}. Deep learning could offer solutions to several computational tasks important to the industry, e.g., B-Rep generation and solid segmentation. Aiming to solve these tasks, various works have been devoted to learning the geometry and topology of B-Rep data~\cite{jayaraman2021uvnet,cadnet2022,jayaraman2023solidgen,willis2022joinable, xu2024brepgen}. A major work, BrepGen~\cite{xu2024brepgen}, decomposes the B-Rep entities (faces, edges, vertices) into a tree data structure that encodes the topological information. 

Despite the success of learning the topology of B-Rep, we find that these works~\cite{xu2024brepgen, jayaraman2021uvnet, meltzer2021uvstyle} utilize a parametric approximation to surfaces, that is representing the surface with a certain number of 3D points uniformly distributed in the UV-domain, \ie, a UV-grid~\cite{jayaraman2021uvnet}. More details about UV-grids are introduced in~\cref{sec:preliminaries}. However, this approach has several drawbacks: (1) the approximation to the target surface is often imprecise unless a dense UV-grid is used to achieve an acceptable error range~\cite{jayaraman2021uvnet, Nurbs_fit_occ_2018, Nurbs_fit_occ_2019}. (2) The demand for high accuracy often incurs large data sizes, model sizes, and computational costs, which is sometimes unnecessary. For example, planar surfaces are represented by 32$\times$32 3D points in~\cite{xu2024brepgen}. (3) Generative models that use UV-grid-based surface approximation, e.g., BrepGen, can produce artificial undulating patterns on the surface. For instance, we showcase several solids generated by BrepGen in~\cref{fig:Nurbs-Gen_qualitative_zoom} (a), where some sections of the curvy surface are not perfectly smooth.

\paragraph{Motivation} Alternatively, it is more natural and advantageous to use the NURBS parametrization in the solids learning task: NURBS are more accurate~\cite{piegl_nurbs_1996, Nurbs_fit_occ_2019,nurbs_diff_2022}, easier to manipulate~\cite{NurbsSurvy, hoschek_fundamentals_1996}, and have less parameters~\cite{piegl_nurbs_1996, handbook_cad} compared to UV-grids. However, incorporating the parameters of a NURBS surface in deep neural networks is not trivial: (1) The parameters --- control points, knot vectors, and weight matrix (see explanation in~\cref{sec:preliminaries}) --- have varying sizes across surfaces in a B-Rep data set. It is challenging to unify them into a fixed-size input to a deep neural network. (2) These parameters are related (e.g., control points are only meaningful together with knot vectors and and the dimension of the weight matrix depends on the number of control points), and hence, they should be encoded and decoded jointly if we wish to learn a shared latent representation thereof. 

\paragraph{Contribution} We target learning \emph{an effective representation of NURBS parameters}, for which we propose the \emph{\modelname pipeline}\footnote{Code can be found under: \url{https://github.com/jiajie96/NeuroNURBS-python.git}} consisting of preprocessing and autoencoding. We utilize the transformer architecture~\cite{attention2017, bert_2019} for the autoencoder to handle the heterogeneous sizes of NURBS parameters. We refer to the resulting latent representation as \emph{NURBS features}. More precisely, \emph{\modelname} is our proposed pipeline to convert \emph{NURBS parameters} into \emph{NURBS features} that can be used in downstream tasks, e.g., solid segmentation and generation. In~\cref{fig:Nurbs-Gen_qualitative_zoom} (b), we illustrate some examples of surfaces generated with NURBS-Gen, which is obtained by replacing the preprocessing and autoencoder of surface entities with \modelname in the BrepGen~\cite{xu2024brepgen} framework. As seen from the figure, the surfaces generated with \modelname and NURBS features are very smooth and regular. To evaluate \modelname, we conduct the following experiments:
\begin{enumerate}
    \item We compare its performance with the UV-grid method for reconstructing surfaces on DeepCAD data~\cite{Wu_2021_deepcad} in terms of accuracy, memory efficiency, and computation cost (\cref{sec:neuroNurbs_efficiency}). We see the memory required to store features of solids is reduced by \textbf{79.9\%}, the surface autoencoder's size is reduced by \textbf{92.9\%}, the GPU consumption for training the surface autoencoder is reduced by \textbf{86.7\%}.
    
    \item We test the \modelname on two downstream tasks: B-Rep generation on DeepCAD~\cite{Wu_2021_deepcad} (\cref{sec:neuroNurbs_generation}) and  solid segmentation on MFCAD~\cite{mfcad2020} (\cref{sec:neuroNurbs_segmentation}). For the former, our NURBS-Gen improves the FID (Fr\'echet Inception Distance~\cite{heusel2017ttur}) from 30.04 (achieved by BrepGen) to \textbf{27.24}. For the latter, we utilize a Graph Attention Network (GAT)~\cite{2018gat}, resulting in a new model called NURBS-GAT. It achieves an accuracy of \textbf{99.65\%}.
\end{enumerate}   

\section{Related Work}
\label{sec:relatedwork}

Learning from 2D geometric designs~\cite{saalae_fan_2024, machines11080802, ELYAN202091,fan2023noiseschedulinggeneratingplausible} is not sufficient for the rapidly evolving industry, where designs are mainly represented in 3D Boundary Representation (B-Rep)~\cite{heidari2024geometric} in computer-aided design (CAD). Huge progress has been made in using modern neural networks to tackle 3D learning tasks: representing them in point cloud~\cite{qi2016pointnet} or mesh~\cite{hanocka2019meshcnn} enables the learning on 3D solids and achieves first success in solid learning; further researches leverage the generative power of Diffusion Models~\cite{ho2020ddpm, song2022ddim} to enable the generation of point clouds~\cite{Zhou_2021_ICCV,zeng2022lion}, meshes~\cite{liu2023meshdiffusion, hui2022neural}, and voxels~\cite{li2023diffusion}. However, learning directly on B-Rep shapes is still challenging, and some works, e.g., DeepCAD for B-Rep generation~\cite{xu2024brepgen}, get around this by representing the CAD shape as a sequence of modeling operations (e.g., sketching and extruding) stored in parametric CAD files~\cite{step_based_learning_2023, Wu_2021_deepcad,Fusion_360_gallery_2021,Lambourne_2022, xu2022skexgen}, where modeling operations are limited to the sketching and extruding and are difficult to scale~\cite{jayaraman2023solidgen}.

To directly consume the geometric and topological information from B-Rep data, recent attempts convert the B-Rep into a graph, then pass through a graph neural network (GNN), hereby performing learning tasks such as face segmentation or solid classification~\cite{Brep2graph_1985,jayaraman2021uvnet, cadnet2022}. Prediction accuracy has been significantly improved by increasing the information added to the graph, i.e., from using face normal and distance to origin as node feature in CADNet~\cite{cadnet2022} to additionally introducing UV-grids, trimming mask and curve geometries into the graph in UV-Net~\cite{jayaraman2021uvnet}. While converting B-Rep to graphs performs well in shape classification and segmentation tasks, in the field of solid generation, B-Rep is often represented as a predefined hierarchical structure and generated by autoregressive prediction of B-Rep entities. This approach has given rise to some convincing methods, such as SolidGen~\cite{jayaraman2023solidgen} and BrepGen~\cite{xu2024brepgen}.

It is worth noting that current research has hardly attempted geometric learning using Non-Uniform Rational B-Splines (NURBS)~\cite{piegl_nurbs_1996}, but the machine learning community has already developed a keen interest in learning from NURBS. For example, fitting NURBS surfaces from point clouds with gradient-based learning~\cite{nurbs_diff_2022, elmidany2011nurbs} has been well studied, where the challenge of reversing point clouds to NURBS surfaces have been pointed out and more work has been done to solve this problem~\cite{Nurbs_fit_occ_2018, Nurbs_fit_occ_2019}. But using point cloud (UV-grid) as surface representation is still suboptimal.

\section{Method}
UV-grids enable the direct learning on B-Rep solids, but a UV-grid still serves as an approximation to the source surface, whereas using NURBS parametrization is a more natural and advantageous choice: NURBS are more accurate~\cite{piegl_nurbs_1996, Nurbs_fit_occ_2019,nurbs_diff_2022} and have less parameters~\cite{piegl_nurbs_1996, handbook_cad}. However, using modern neural networks to operate NURBS parameters is understudied and remains challenging (see~\cref{sec:nurbs_embedding}). To address this, in this section, we describe \modelname\, which consists of a preprocessing pipeline for NURBS parameters and an autoencoder that is able to encode NURBS parameters into a low-dimensional feature space for downstream application.

\subsection{Preliminaries}
\label{sec:preliminaries}
\paragraph{UV-grids}
A 3D freeform surface $\bm{S}$ can be approximated by a parameterized function $\mathbf{F}(u, v)\colon \R^2 \rightarrow \R^3$, which is learned on a number of points evenly distributed in $[\umin,\umax]\times [\vmin,\vmax] \in \mathbb{R}^2$. $\mathbf{F}(u, v)$ is called UV-grid function (see the left side of~\cref{fig:NurbsEmb-Diagram}). To learn the grid function, one can sample $n \times m$ points (along the $U$ and $V$ dimensions, respectively) from the target surface. A large sample number can increase the approximation accuracy of the original surface but this comes at the cost of higher-dimensional representation, leading to increased training times and memory requirements.

\paragraph{NURBS}
Non-Uniform Rational B-Splines (NURBS) are essentially B-splines applied in homogeneous coordinates, rather than 3D coordinates, i.e., $(x, y, z, w) \mapsto (x/w, y/w, z/w)$. Each NURBS of order $n$ (polynomial of degree $n-1$) requires $n$ control points and $2(n-1)$ knots. A NURBS curve $C(u)$ can be defined using coordinate $u$ with the formula: 
\begin{equation}
C(u) = \frac{\sum_{i=1}^n N_i^p(u)w_i \bm{p}_i}{\sum_{i=1}^n N_i^p(u)w_i} \; ,
\end{equation}
where $\bm{p}_i$s are 3D control points and $N_i^p$s are basis functions of degree $p \leq n-1$, defined recursively on a knot vector $(u_1, \ldots, u_i, \ldots, u_{n+p+1})$:
\begin{align}
&N_i^p(u)=\frac{u-u_i}{u_{i+p}-u_i}N_i^{p-1}(u)+\frac{u_{i+p+1}-u}{u_{i+p+1}-u_{i+1}}N_{i+1}^{p-1}(u) \label{eq:NURBSBasis} \\
&N_i^0(u) = 
\begin{cases}
    1 & \mbox{if $u_i\leq u < u_{i+1}$} \; , \\
    0 & \mbox{otherwise}.
\end{cases} \label{eq:BSplineBasis2}
\end{align}
A NURBS surface $\bm{S}(u,v) \colon \R^2 \rightarrow \R^3$ is the tensor product of two NURBS curves, $C_U(u)$ (on $n$ control points with order $p$) and $C_V(v)$ (on $m$ control points with order $q$):
\begin{align}
    &\bm{S}(u, v) \coloneqq C_U(u)C_V(v) \nonumber \\
    &= \frac{\sum_{i=1}^n{\sum_{j=1}^m{N_i^p(u) N_j^q(v)w_{ij}\bm{P}_{ij}}}}{\sum_{i=1}^n{\sum_{j=1}^m{N_i^p(u) N_j^q(v)w_{ij}}}}, \label{eq:NURBS}
\end{align}
which takes the following parameters: a grid of control points $\bm{P}\in\mathbb{R}^{n \times m \times 3}$, weights $\bm{W} \in \mathbb{R}^{n \times m}$,  the $U$-direction knot vector \(\bm{U}=(u_1, \ldots, u_{n+p+1}) \in \mathbb{R}^{n+p+1}\), and the $V$-direction knot vector $\bm{V}\ =\ (v_1, \ldots, v_{n+q+1}) \in \mathbb{R}^{n+q+1}$.

\subsection{NeuroNURBS}
\label{sec:nurbs_embedding}
To represent a surface, we propose to take its NURBS parametrization directly, which can be extracted from the B-Rep model, e.g., with \texttt{OpenCascade} functionalities. Compared to the UV-grid approach, the NURBS parametrization is (1) smooth and describes the surface precisely; (2) it can be memory efficient since it only requires storing a collection of 3D control points, a weight matrix, and two knot vectors. However, developing a deep learning model that takes as input the parameter of NURBS is challenging: 
\begin{itemize}
    \item NURBS parameters have different dimensions, e.g., a knot vector is a 1D object and control points are 3D objects and vary in size from one surface to the other.
    \item for a generative task, NURBS parameters have to be combined in the encoder and the decoder must map each latent point to a set of valid parameters.
\end{itemize}
We propose \modelname to resolve these challenges. It consists of two components: a preprocessing pipeline for the NURBS parameters (see below) and an autoencoder to learn an effective representation of NURBS parameters. We depict our method in~\cref{fig:NurbsEmb-Diagram}.

\begin{table*}[ht]
\small
\centering
\caption{Performance comparison of UV-grids and NURBS representations for surface reconstruction task. We sample 10k surfaces and solids randomly from the DeepCAD dataset and check the memory needed to store the UV-grids (of size $32\times 32$) and NURBS parameters of the surfaces (input data size). The surface VAE (from BrepGen) taking UV-grids as input has a lot more parameters than the VAE model used in \modelname (see~\cref{subsec:NURBS-VAE}). Also, the distribution learning column shows the differences between the distribution of surfaces in the test set and the ones reconstructed by surface VAE or \modelname. Construction speed compares the time each method takes to convert the surface representation to a NURBS surface. 
}
\renewcommand{\arraystretch}{1.5}
\resizebox{\textwidth}{!}{
\begin{tabular}{llccccccc}
\toprule
& \multicolumn{2}{c}{Input data size}          &\multicolumn{2}{c}{Autoencoder size}         &  \multicolumn{1}{c}{Surface reconstruction} & \multicolumn{2}{c}{Distribution learning} & Construction speed\\
\cmidrule{2-9}
 &Surface      &Solid          &VAE    &GPU     &  MMD                      &  JSD                     &  Coverage   &  speed\\
 &(MB/10k)          &(GB/10k)            &\#Param. &(GB)   &  $(\times -10^3)\downarrow$ &$(\times -10^2)\downarrow$ & $(\%)\uparrow$ & (NURBS/s)$\uparrow$\\
\midrule
UV-grids        & 245.8         &37.7          &84M  & 17.61       &  $\textbf{0.338}$  &$1.26$  &$62.6$ &230\\

NURBS      &  \textbf{8.16 (\textcolor{red}{96.7$\%\downarrow$})} & \textbf{7.6 (\textcolor{red}{79.9\%$\downarrow$})} & \textbf{6M (\textcolor{red}{92.9\%$\downarrow$})} & \textbf{2.35 (\textcolor{red}{86.7\%$\downarrow$})}& $0.552$           &$\textbf{0.87}$  &$\textbf{68.2}$ &\textbf{3230 (\textcolor{red}{92.9\%$\uparrow$})}\\
\bottomrule
\end{tabular}}
\label{tab:neuroNurbs_efficiency}
\end{table*}

\subsubsection{Reading NURBS parameters} \label{subsec:reading}
In a B-Rep solid, the surfaces usually exist in a form of trimmed NURBS surface. To read the NURBS parameters, i.e., control points, knot vectors, we design the following pipeline with functions from \texttt{pythonOCC}~\cite{pythonOCC}: 

\begin{enumerate}[label=\arabic*.]
    \item Convert the surface of form \texttt{TopoDS\_Face} into an untrimmed NURBS surface with \texttt{BRepBuilderAPI\_NurbsConvert()}.
    \item Then, use \texttt{BRep\_Tool.Surface()} and \texttt{geomconvert.SurfaceToBSplineSurface()} of \texttt{geomconvert} to convert the NURBS surface into a \texttt{Geom\_BSplineSurface},
    \item Finally, reading the control points, weights, U- and V-knot vector with \texttt{Pole()}, \texttt{Weight()}, \texttt{UKnotSequence()} and \texttt{VKnotSequence()}.
\end{enumerate}

\subsubsection{Preprocess NURBS parameters} \label{subsec:normalization}
We consider the \textbf{control points} $\bm{P}$, \textbf{weights} $\bm{W}$, and \textbf{knot vectors} $\bm{U}$ and $\bm{V}$ for the learning task, which we call the NURBS parameters. We exclude degrees $p$ (degree in the U-direction) and $q$ (in the V-direction) from parameters since they can be easily calculated: $p$ is the length of knot vector $\bm{U}$ minus  $n+1$, where $n$ is the number of control points in the U direction. The degree $q$ can be determined in the same way.

\paragraph{Normalization} For control points $\bm{P}_{ij} = (x_{ij},y_{ij},z_{ij})$, we determine the coordinate axis with the largest range among three coordinates:
\begin{equation}
    d = \max{(x_{\text{max}}-x_{\text{min}}, y_{\text{max}}-y_{\text{min}}, z_{\text{max}}-z_{\text{min}})},
\end{equation}
and then normalize the coordinates as $\hat{x}_{ij} = (x_{ij}-x_{\text{min}})/d$ (same for y-axis and z-axis).
The normalization ensures $(\hat{x}_{ij}, \hat{y}_{ij}, \hat{z}_{ij}) \in [0, 1]$. Note that weights $w_{ij}$ also take values in the unit interval. Hence, we concatenate $\bm{P}$ and $\bm{W}$ together: $\bm{P_w} = [(\hat{x}_{ij}, \hat{y}_{ij}, \hat{z}_{ij}, w_{ij})]_{i\in[1..n], j\in[1..m]} \in \mathbb{R}^{n \times m \times 4}$.

\paragraph{Padding} With a padding value of $0$, we augment $\bm{P_w}$ to a tensor of size $(d,\,d,\,4)$, where $d$ is the largest number of controls points across all surfaces in a data set. We also pad the knot vectors $\bm{U}$ and $\bm{V}$ to length $k$ (the length of the longest knot vector in a data set). To combine the knot vector and control points, we duplicate the knot vector to a tensor of shape $(d,\, d,\, k)$. Also, we save the padding mask of $\bm{P_w}$ for the autoencoder. Although the padding seems to impair the efficiency of the NURBS representation, the padding parts are omitted when training the transformer-based autoencoder using the padding mask.

\subsubsection{Learning NURBS features with autoencoder}  \label{subsec:NURBS-VAE}
We study recent works in multi-modal and multi-task autoencoding~\cite{ngiam2011multimodal,eigen2015predicting,bachmann2022multimae,palumbo2023mmvae} and design a VAE (variational autoencoder) model to learn a latent representation of the NURBS parameters, which we call \emph{NURBS features}.

Our autoencoder is inspired by MultiMAE~\cite{bachmann2022multimae}. We use separate dense layers for each preprocessed NURBS parameter (control points with weights $\bm{P_w}$, U-knot vector $\bm{U}$ and V-knot vector $\bm{V}$). The sum of the dense outputs is then processed with a sine-cosine position embedding and is then input into a transformer together with the control point padding mask. For the selection of a different transformer backbone from MultiMAE~\cite{bachmann2022multimae}, we do not use the Vision Transformer (ViT)~\cite{dosovitskiy2020ViT} as our data has a low dimension and does not require patch embedding, but the introduction of a padding mask would be helpful. Thus, we select a transformer backbone from BERT~\cite{bert_2019}.
Each transformer block is eight layers of four-head self-attention blocks. Followed up with a fully connected dense layer, we obtain $\mu \in \mathbb{R}^{d_z}$ and $\sigma \in \mathbb{R}^{d_z}$, which will then be reparameterized into the surface token $Z \in \mathbb{R}^{d_z}$, like every VAE model. To reconstruct the NURBS parameters from the NURBS feature $Z$, we employ multi-task decoding, namely a separate decoder for each parameter. See~\cref{fig:NurbsEmb-Diagram} for the visualization of the model architecture. 

During the training, we compute the loss: \(  L = \sum_{i=1}^{3}\|\mathbf{x}_i - \mathbf{x}^\prime_i\|^2_2,\)
where $\mathbf{x}_1 = \mathbf{P}_w$, $\mathbf{x}_2 = \mathbf{U}$, $\mathbf{x}_3 = \mathbf{V}$ are the NURBS parameters, and $\mathbf{x}_1^\prime = \mathbf{P}_w^\prime$, $\mathbf{x}_2^\prime = \mathbf{U}^\prime$, $\mathbf{x}_3^\prime = \mathbf{V}^\prime$ are the reconstructed NURBS parameters. Practically, adding a KL-divergence term could help convergency. For the VAE training and inference, inputs are normalized to [$-1, 1$] and outputs of decoding will be denormalized to [$0, 1$]. Notably, for generated NURBS parameters, we perform the following post-generation checks: we remove control points with weight less than or equal to $0$, and clip knot vectors once the value in the sequence reaches $1$.
\section{Experiments}

We performed several experiments to investigate the effectiveness of our \modelname. NURBS parameters are more efficient than UV-grids for surface representation and can also be used as learning data. To demonstrate this, we compare its performance with UV-grids w.r.t.~accuracy, memory efficiency, and computation cost (\cref{sec:neuroNurbs_efficiency}) on the DeepCAD dataset. After showing NURBS's utility in efficient representation, we introduce NURBS features encoded by \modelname into two downstream tasks: (1) The CAD (B-Rep) generation task on the DeepCAD dataset(\cref{sec:neuroNurbs_generation}), and (2) The solid segmentation task on the MFCAD data set (\cref{sec:neuroNurbs_segmentation}).
\subsection{Datasets}
\label{sec:dataset}

\paragraph{DeepCAD}
We introduce the DeepCAD~\cite{Wu_2021_deepcad} dataset for the experiment of surface reconstruction and solid generation. The DeepCAD data set originally contains CAD data in the form of a sequence of engineering operations that can be converted into B-Reps with their pre-defined construction algorithm. We follow the previous work~\cite{xu2024brepgen}, i.e., filtering out invalid B-Reps shapes (in case of multiple solids or being not watertight) and B-Reps with more than 30 surfaces. We use the train-val-test splitting of DeepCAD~\cite{Wu_2021_deepcad}, resulting in 69\,512 B-Reps for training and 7\,083 B-Reps for testing. Closed surfaces, e.g., cone surfaces, are cut into two separate NURBS surfaces as done in~\cite{jayaraman2023solidgen}. In representing surfaces as UV-grids, we set $n=m=32$, the same as~\cite{xu2024brepgen}. For training the autoencoder in \modelname, we list all NURBS surfaces from the training solids and remove duplicated surfaces, obtaining 59\,961 unique NURBS surfaces. The preprocessing of NURBS surfaces has been detailed in~\cref{sec:nurbs_embedding}, and for the DeepCAD dataset, we set the \modelname\ hyperparameters $d=10, k=10, d_z=48$. The DeepCAD dataset is implemented for evaluating the efficiency in data representation~\cref{sec:nurbs_embedding} and B-Rep solid generation~\cref{sec:neuroNurbs_generation}.

\paragraph{MFCAD}
For the solid segmentation experiment in~\cref{sec:neuroNurbs_segmentation}, we introduce the MFCAD~\cite{mfcad2020} dataset of 15\,488 CAD solids in the STEP file format, which is the standard file format for B-Rep. MFCAD is designed for the machining feature segmentation task, where there are 16 various face labels and each face is labeled with a certain class. We split the MFCAD solids into train-val-test datasets with a ratio of 70-15-15, resulting in 10\,842 solids for training. For all surfaces from training solids, we first remove duplicated surfaces and then split the surfaces into train-val-test datasets with a ratio of 90-5-5, where we collect 24\,603 NURBS surfaces for training the \modelname. The surface embedding follows the pipeline described in~\cref{sec:nurbs_embedding}, where we set $d=2, k=4, d_z=48$. We take $d=2$ since MFCAD contains only planar surfaces, which requires minimally four control points to describe with NURBS.

\subsection{Surface reconstruction}
\label{sec:neuroNurbs_efficiency}
On the DeepCAD~\cite{Wu_2021_deepcad} dataset, we compare our method with the UV-grid method for reconstructing the surface. For the UV-grid representation method, we take the surface VAE (Variational AutoEncoder) from BrepGen~\cite{xu2024brepgen}. The goal is to test our \modelname (the VAE part) proposed in~\cref{subsec:NURBS-VAE} (which takes as input NURBS parameters) against their surface VAE (which takes UV-grid data), in terms of input data size, the model size, the reconstruction error, distribution learning, and NURBS construction speed. We summarize the results in~\cref{tab:neuroNurbs_efficiency}. 

\paragraph{Input data size}
We randomly sample 10k surfaces from the DeepCAD dataset and generate UV-grids data on a $32\times 32$ grid for each surface. We also collect the NURBS parameters for these surfaces. We see from~\cref{tab:neuroNurbs_efficiency} that the NURBS parameters reduce the memory cost by 96.7\% compared to the UV-grid data for the surfaces. For the representation of solids, we follow the solid preprocessing in BrepGen~\cite{xu2024brepgen} to collect the learning data, where the B-Rep entities are represented as a tree data structure, and the surfaces are represented in either UV-grids or NURBS. Here, we observe that the memory cost drops by 79.9\% when the surfaces are represented using NURBS.

\paragraph{Computational cost, reconstruction error, and distribution learning}
The model size and computational cost of the VAE also differ drastically. In~\cref{tab:neuroNurbs_efficiency}, we observe the surface VAE from BrepGen has 84M parameters, which takes a UV-grid of dimension $32 \times 32$. In contrast, our method with setting $d=10, k=10, d_z=48$ has only \textbf{6M} parameters. Also, we measure the GPU consumption for training both VAE models with batch size $512$: the surface VAE for UV-grids requires $17.61$GB GPU memory, while our method requires only $2.35$GB --- an \textbf{86.7\%} reduction.

For the reconstruction error, we measure, between input and reconstructed surfaces (10k each), the Minimum Matching Distance (MMD) metric on 2\,000 points sampled from the surface. MMD measures the mean Chamfer Distance between a sample in the test dataset and its closest sample in the generated ones. The Chamfer Distance (CD) is a commonly used metric to measure the similarity between two point clouds. The mean and standard error of MMD is reported over 10k surfaces in~\cref{tab:neuroNurbs_efficiency}, where \modelname has comparable performance to the surface VAE. 

To check if the VAE can approximate the distribution of surface data (which is important in the downstream generative task), we also measure the Coverage (COV) and Jensen-Shannon Divergence (JSD) metrics on the test set of DeepCAD. Coverage is the percentage of test data that has at least one match after assigning every generated sample to its closet test data based on CD. JSD is commonly used to measure the distribution distance between the test and generated data. The results show that \modelname is consistently closer to the distribution of surfaces on the test data.

\paragraph{NURBS construction}
For the B-Rep generation task, if one exploits the UV-grid representation of a surface, the UV-grid must be converted to a NURBS surface with surface fitting functions, e.g., \texttt{GeomAPI\_PointsToBSplineSurface} function of \texttt{PythonOCC}. In contrast, if one uses \modelname, the only step needed is to revert the normalization of NURBS parameters (see~\cref{subsec:normalization}) and feed the resulting parameters to a CAD tool, e.g., \texttt{Geom\_BSplineSurface} function in \texttt{PythonOCC}, which is computationally very cheap. From~\ref{tab:neuroNurbs_efficiency}, we see that our method increases the construction speed of NURBS by \textbf{92.9\%}.

More importantly, we delve into the degrees of the reconstructed NURBS surface by both methods and check whether the degree of the reconstructed ones agrees with the surfaces in the test dataset. In~\cref{tab:degree_inheritance}, we show the distribution of NURBS degrees for the surfaces in the test data, for the surface VAE and \modelname. We observe that for reconstructing the test surfaces, our method produces a distribution of NURBS degrees which perfectly matches the test data since our method directly learns a latent representation of the NURBS parameter. However, the UV-grid method concentrates on higher degrees when fitting NURBS surfaces to the UV-grid. We argue that this could be the main reason for the undulating issue shown in~\cref{fig:Nurbs-Gen_qualitative_zoom} as higher-order polynomials often introduce artificial fluctuations.

\begin{table}
\small
\centering
\caption{The empirical distribution of the degree of polynomials along U(V) direction are measured in two tasks: surface reconstruction and generation. For the former, we train the UV-grid-based VAE of \cite{xu2024brepgen} and our \modelname on the training set of DeepCAD and reconstruct all surfaces in the test set. For the latter, we employ BrepGen~\cite{xu2024brepgen} and our NURBS-Gen to create new solids and measure if the surface degree distribution of the generated solids agrees with that of the test solids. Notably, one must perform a NURBS fitting to the surface approximation obtained from UV-grid VAE. 
}
\resizebox{\linewidth}{!}{
\begin{tabular}{llccc}
\toprule
\multirow{2}{1.5cm}{NURBS source}&  \multicolumn{4}{c}{Frequency of U (V) degrees}\\
 \cmidrule{2-5}
     &$2$  & $3$ & $4$ &$\geq5$\\
\midrule
\multicolumn{5}{l}{Surface reconstruction}\\
\midrule
Test data (surfaces)     &88.59\%   &2.03\%    &0.00\% &9.38\%  \\
UV-grid VAE~\cite{xu2024brepgen}        &0.00\%    &94.01\%   &5.92\% &0.07\% \\
\modelname     &\textbf{88.59\%}   &\textbf{2.03\%}    &\textbf{0.00\%} &\textbf{9.38\%}  \\
\midrule
\multicolumn{5}{l}{Surface generation}\\
\midrule
Test data (solids)    &93.45\%   &2.32\%    &0.00\% &4.50\%  \\
BrepGen~\cite{xu2024brepgen}        &0.00\%    &0.00\%   &99.99\% &0.01\% \\
NURBS-Gen        &88.69\%    &0.88\%   &0.00\% &10.43\% \\
\bottomrule
\end{tabular}}
\label{tab:degree_inheritance}
\end{table}

\subsection{CAD (B-Rep) generation}
\label{sec:neuroNurbs_generation}

\begin{figure*}[th]
  \centering   
\begin{subfigure}[b]{0.31\linewidth}
    \centering
    \includegraphics[width=1\linewidth, trim=0mm 0mm 0mm 0mm, clip]{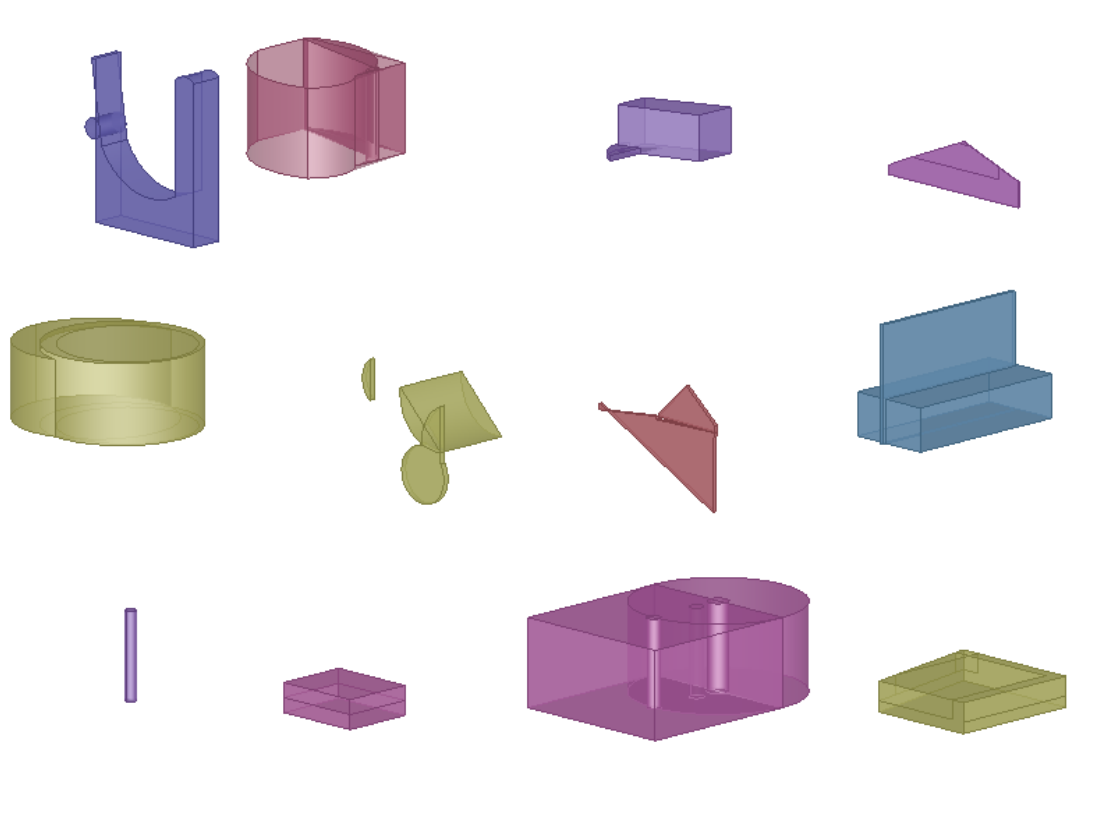}
    \caption{DeepCAD}
\end{subfigure}
\hfill
\unskip\ \vrule\ 
\hfill
\begin{subfigure}[b]{0.31\linewidth}
    \centering
    \includegraphics[width=1\linewidth, trim=0mm 0mm 0mm 0mm, clip]{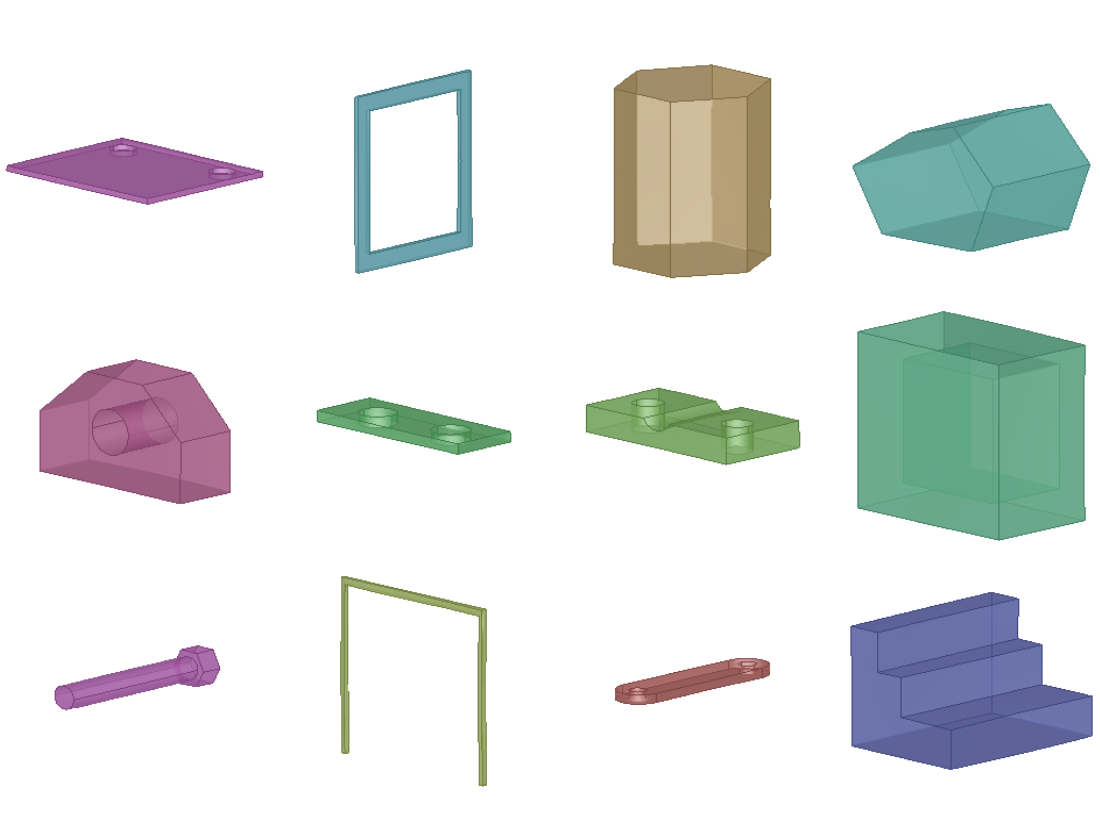}
    \caption{BrepGen}
\end{subfigure}
\hfill
\unskip\ \vrule\ 
\hfill
\begin{subfigure}[b]{0.31\linewidth}
    \centering
    \includegraphics[width=1\linewidth, trim=0mm 0mm 0mm 0mm, clip]{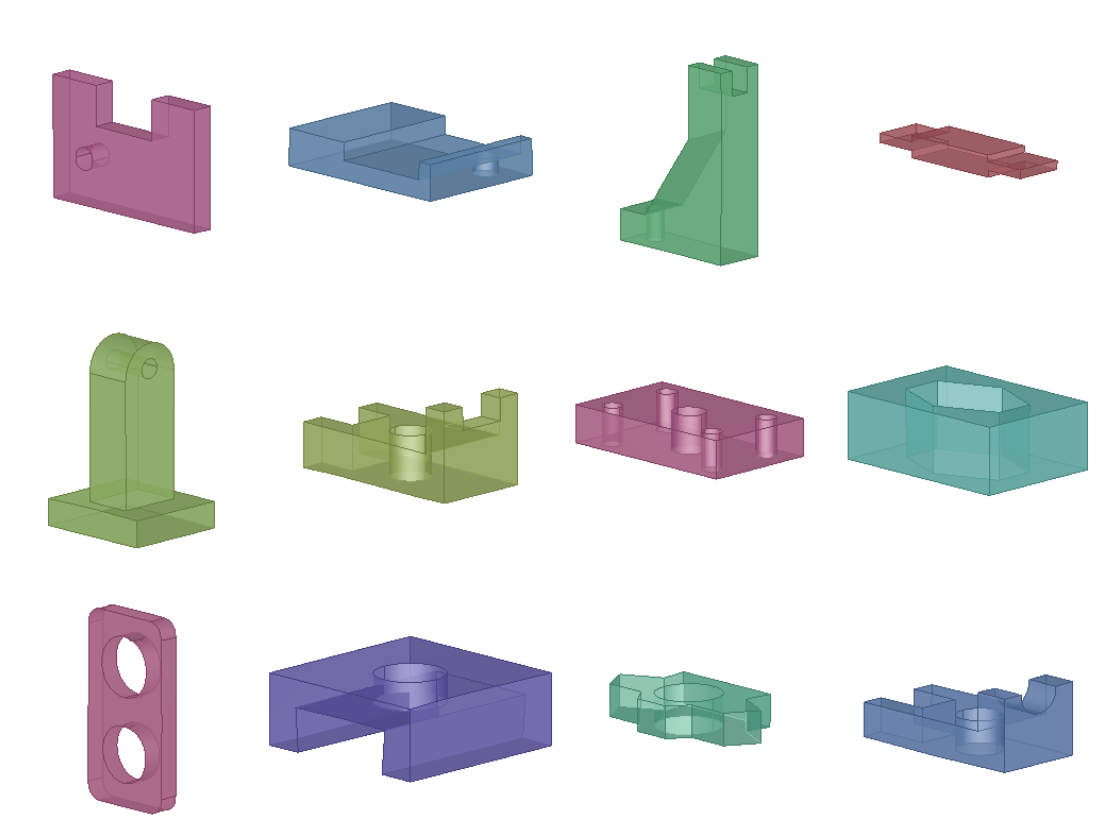}
    \caption{NURBS-Gen}
\end{subfigure}
\caption{Qualitative evaluation. It is observe that DeepCAD often generates invalid (i.e., multiple and collapsed) solids; while BrepGen and our NURBS-Gen are able to synthesize clean and upstanding B-Rep solids. In~\cref{fig:Nurbs-Gen_qualitative_zoom}, we visualize generated samples from BrepGen and NURBS-Gen with zooming-in for a deep-diving qualitative evaluation.
}
\label{fig:Nurbs-Gen_qualitative}
\end{figure*}

\begin{figure*}[t]
    \centering
    \includegraphics[width=1\linewidth, trim=5mm 3mm 4mm 8mm, clip]{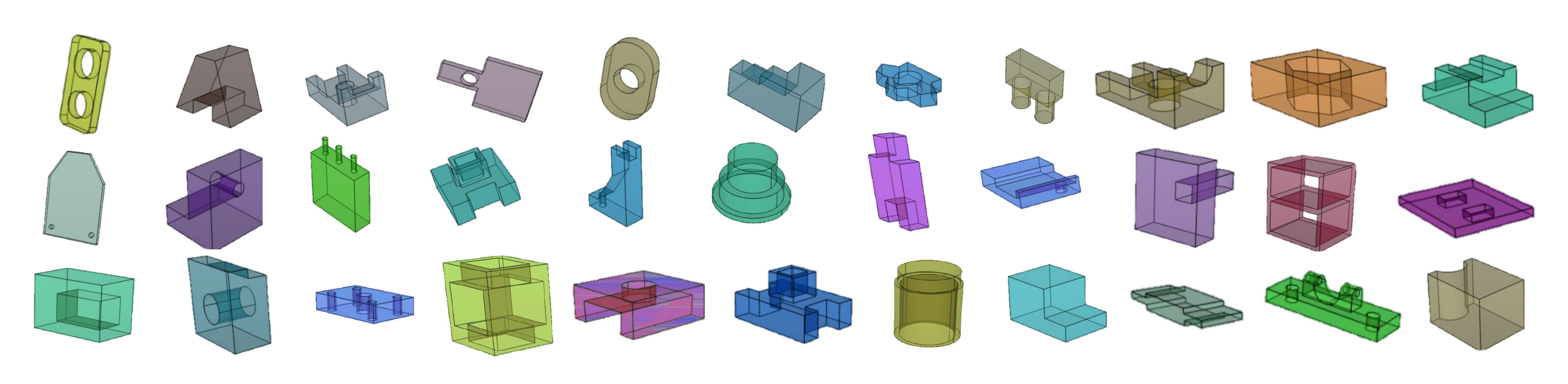}
    \caption{More examples generated with NURBS-Gen.}
    \label{fig:Nurbs-Gen_qualitative_extra}
\end{figure*}

BrepGen~\cite{xu2024brepgen} generates B-Rep solids by progressively creating the B-Rep entities (i.e., faces, edges, and vertices) following a hierarchical tree structure from top to bottom with four diffusion models, which are Transformer-based DDPM~\cite{ho2020denoising}. For generating faces and edges, each node in the tree structure contains a node feature representing the global bounding box of each face/edge and a latent code for local geometry, e.g., the shape of the surface. BrepGen represents all B-Rep surfaces as UV-grids, and fits UV-grids to NURBS surfaces after generation. Here, we substitute the surface VAE model, implemented by BrepGen for autoencoding UV-grids, with our \modelname~(\cref{subsec:NURBS-VAE}). Technically, we set the latent dimension of our NURBS VAE to $d_z=48$, which is exactly the same as the latent dimension of surface VAE in BrepGen. Then, we fine-tune the four pre-trained diffusion models in BrepGen (i.e., diffusion models for face bounding box, face geometry, edge bounding box, edge geometry, and vertex coordinates) on the DeepCAD data set. We call the resulting model NURBS-Gen. As for the details, face denoisers are fine-tuned with a batch size of 256 for 1\,000 epochs, the edge denoisers with a batch size of 64 for 500 epochs, and the training is conducted using one NVIDIA A10G GPU. As for the inference, the sampling process of NURBS-Gen follows the pipeline of BrepGen and the surface decoding is replaced by the decoder in \modelname. 

\paragraph{Results} We generate 3\,000 B-Rep solids with DeepCAD model~\cite{Wu_2021_deepcad}, BrepGen~\cite{xu2024brepgen}, and our NURBS-Gen and compare them to 1\,000 test solids of the DeepCAD Dataset. For the DeepCAD model and BrepGen, we utilize their model checkpoint that is pre-trained on the DeepCAD dataset. Between the generated and test solids, we compute Minimum Matching Distance (MMD), Coverage (COV), and Jensen-Shannon Divergence (JSD) metrics. To calculate these metrics, we convert each solid into a point cloud with 2\,000 points, following the evaluation method in~\cite{Wu_2021_deepcad,xu2024brepgen}. We show the results in~\cref{tab:nurbsgen_quantitative}: NURBS-Gen is quite comparable with BrepGen w.r.t. MMD and COV, and slightly better in JSD values.

We argue that the above evaluation method, which takes as input point clouds, might be biased toward methods like BrepGen, which utilizes UV grids. Hence, we additionally measure the Fr\'echet Inception Distance (FID)~\cite{heusel2017ttur} between the 3D solids rendered in 2D with a certain viewpoint. We repeat the FID calculation for 20 different viewpoints and report an average FID value, as proposed in~\cite{xiong2024octfusion, 3DShape2VecSet2023}. From~\cref{tab:nurbsgen_quantitative}, our NURBS-Gen has a better FID value compared to BrepGen. 

To check the validity of the generated solids, we first convert each generated B-Rep solid to a STEP file until 3\,000 STEP files are obtained. Then, we use the \texttt{BRepCheck\_Analyzer()} and \texttt{IsValid()} functions from \texttt{pythonOCC} to test the watertightness of each solid. From this process, we calculate a valid ratio among all generated solids (including the ones not saved as STEP files) for each method. Again, in~\cref{tab:nurbsgen_quantitative}, our method shows a higher valid ratio. Also, we visualize some generated solids from DeepCAD, BrepGen, and NURBS-Gen in~\cref{fig:Nurbs-Gen_qualitative_zoom} and~\cref{fig:Nurbs-Gen_qualitative}. In~\cref{fig:Nurbs-Gen_qualitative_zoom}, we observe that BrepGen solids are undulating, i.e., they exhibit the wobbly geometry described in~\cite{xu2024brepgen}. In contrast, our NURBS-Gen can generate smooth and regular surfaces and accurately join the surfaces.

\begin{table}
\small
\centering
\caption{In the CAD (B-Rep) generation task, we list the performance of NURBS-Gen and other related models in terms of the COV, MMD ($\times\-10^2$), JSD ($\times\-10^2$), and Valid ratio. 
}
\small
\resizebox{\linewidth}{!}{
\begin{tabular}{lccccc}
\toprule
 Model  &  MMD $\downarrow$              &  JSD $\downarrow$              &  COV  $(\%)\uparrow$          &FID $\downarrow$                 &Valid $(\%)\uparrow$\\
\midrule
DeepCAD &  $1.41$           & 3.92              &\textbf{77.9}    &\textbf{14.36}     & 34.79    \\
BrepGen & $\textbf{1.02}$   & 1.16              &74.7             &30.04              & 60.95    \\
NURBS-Gen & $1.10$          & \textbf{0.98}     &73.9             &27.24              & \textbf{64.58}        \\
\bottomrule
\end{tabular}}
\label{tab:nurbsgen_quantitative}
\end{table}

\subsection{Segmentation}
\label{sec:neuroNurbs_segmentation}

\begin{figure}[t]
    \centering
    \includegraphics[width=1\linewidth, trim=5mm 3mm 4mm 8mm, clip]{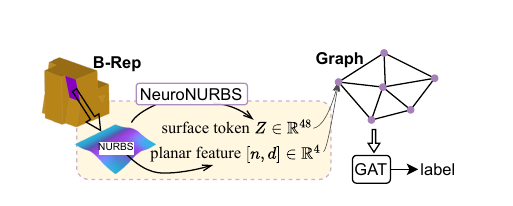}
    \caption{NURBS-GAT diagram.}
    \label{fig:Nurbs-GAT-Diagram}
\end{figure}

\begin{figure*}[t!]
    \centering
    \includegraphics[width=1\linewidth, trim=0mm 1mm 4mm 3mm, clip]{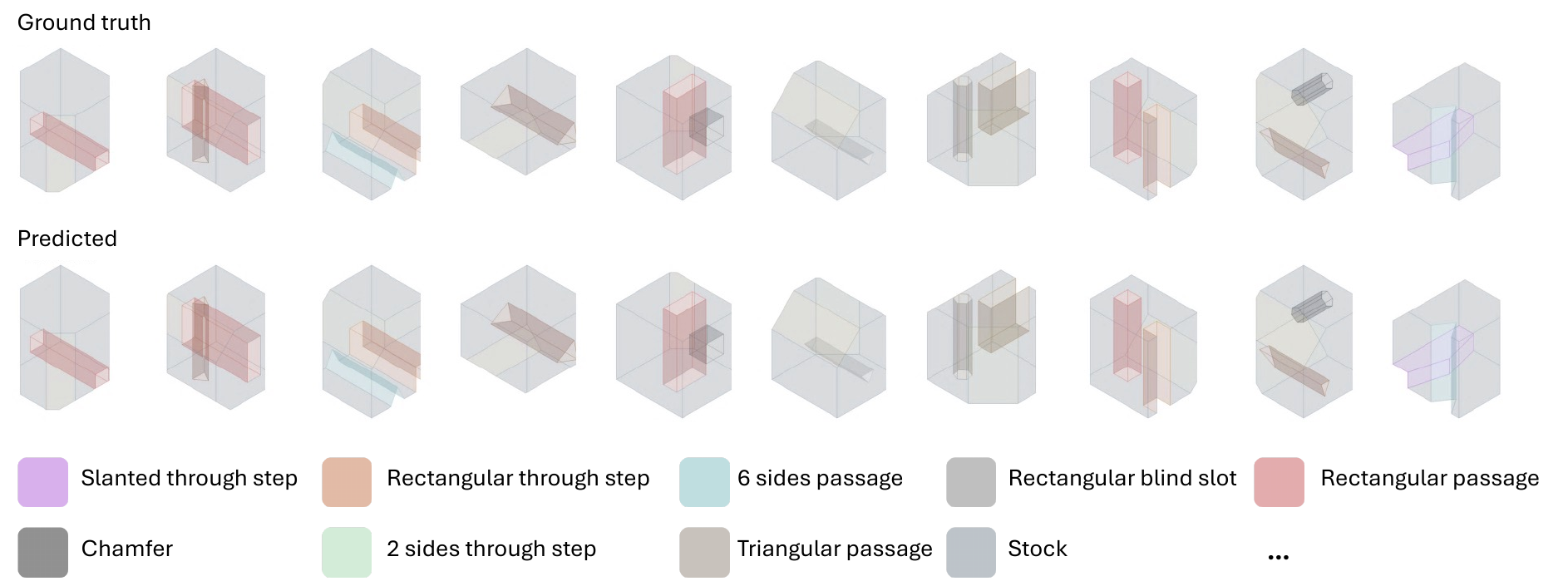}
    \caption{Qualitative evaluation of NURBS-GAT in machining feature segmentation task. Since the per-face accuracy has already achieved 99.65\%, the predicted labels are the same as the ground truth.}
    \label{fig:machining_feature_segmentation_qualitative}
\end{figure*}
In the task of machining feature segmentation~\cite{mfcad2020, cadnet2022}, current state-of-the-art model~\cite{jayaraman2021uvnet} tends to leverage increasing information of geometry or topology (e.g., face normal vector, face geometry, trimming mask, face adjacent or edge features) from the source B-Rep solid to perform accurate machining feature recognition.

We design a NURBS-based segmentation model, NURBS-GAT, that leverages a Graph Attention Network (GAT)~\cite{2018gat} to operate the graph data derived from B-Rep. Replacing the graph convolutional network (GCN) in CADNet with a GAT is due to constant memory issues when increasing the size of the node features. For the graph data, a stack of NURBS feature encoded with NeuroNURBS, face normal vector $\bm{n}$ and coefficient $d$ 
 calculated from a planar equation serves as the $(48+3+1)$-dimensional node feature $V$ and the face adjacency serves as edge connection, as shown in~\cref{fig:Nurbs-GAT-Diagram}. Now, we use the ablation study to show the importance of each input feature by conducting the ablation study on NURBS-GAT.

\paragraph{NURBS-GAT (using NURBS features only)} We remove the face normal vector $\bm{n}$ and coefficient $d$ from the node features of NURBS-GAT, resulting in node feature $V \in \mathbb{R}^{48}$.

\paragraph{CADNet (with GAT)} We remove the NURBS feature from the node of NURBS-GAT, resulting in node feature $V \in \mathbb{R}^{4}$.

\paragraph{UV-GAT} We replace the \modelname\ embedding with UV-grids embedding in node features of NURBS-GAT, resulting in node feature $V \in \mathbb{R}^{52}$. The UV-grids embedding is obtained by training a surface VAE model, same as in~\cref{sec:neuroNurbs_efficiency}, on MFCAD surface training data.

The evaluation in ablation study is conducted by reporting the per-face accuracy in~\cref{tab:neuronurbs_segmentation}, following the usual process in the machining feature segmentation task~\cite{cadnet2022, jayaraman2021uvnet}. Observed from the results, utilizing the \modelname\ embedding is able to improve the per-face accuracy from 92.18\% to 99.65\%, which shows a comparable result to UV-grid embedding. We also conduct a qualitative evaluation of NURBS-GAT in the machining feature segmentation task in~\cref{fig:machining_feature_segmentation_qualitative}, but since the accuracy has already reached 99.65\%, one cannot see any difference between the prediction and the ground truth.

\begin{table}
\small
\centering
\caption{Ablation study on machining feature segmentation task.  Although UV-GAT shows a comparable results to our NURBS-GAT, but their model size is significantly larger.}
\resizebox{\linewidth}{!}{
\begin{tabular}{llcc}
\toprule
Ablated model             &Acc. (\%)  & \#Param. \\
\midrule
CADNet (with GAT)  &92.18          &0.02M  \\
UV-GAT            &99.63           &34.2M \\
NURBS-GAT (only NURBS feature)  &97.03     &1.28M \\
NURBS-GAT         &\textbf{99.65}  &1.29M  \\
\bottomrule
\end{tabular}}
\label{tab:neuronurbs_segmentation}
\end{table}

\section{Conclusion}
With data representation becoming one of the major challenges for solid learning, our work directly operates the native representation of B-Rep surfaces, i.e., NURBS parameters, by proposing \modelname. Our \modelname can embed NURBS parameters into a low-dimension NURBS feature. Using \modelname has demonstrated its ability of significantly reducing storage and GPU consumption, while delivering a comparable performance to state-of-the-art method on both solid generation and segmentation. We also observe that \modelname can remarkably help with the undulation problem in generated surfaces, which could be caused by using an approximate representation (such as UV-grids) to represent source surfaces.

\paragraph{Limitation and future work.} As a first attempt to integrate NURBS parameters into solid learning tasks, there is still much room for improvement in neural networks, e.g., directly learning on NURBS parameters without using an autoencoder. For the solid generation task, the edge geometry in NURBS-Gen is still represented by sampling points in the parametric domain, and it is also technically possible to replace it with NURBS parameters. Moreover, it is a major challenge how to evaluate the generated solids. As we have shown in the quantitative evaluation, DeepCAD seems to perform better in the qualitative evaluation, but its generated solids in visualization are not convincing.
{
    \small
    \bibliographystyle{ieeenat_fullname}
    \bibliography{main}
}


\end{document}